\documentclass[runningheads]{llncs}
\usepackage{graphicx}

\usepackage{csquotes}

\usepackage{wrapfig}
\usepackage{enumitem}
\usepackage{multirow}
\usepackage[belowskip=-6pt,aboveskip=1pt]{caption}
\usepackage{float}
\usepackage{textcomp}
\usepackage[mathscr]{eucal}

\usepackage{changepage}
\usepackage{amsmath,amssymb,amsfonts}
\usepackage{algorithm}
\usepackage{algorithmic}
\usepackage{textcomp}
\usepackage[colorlinks]{hyperref}
\usepackage{booktabs}
\usepackage{soul}
\usepackage{amsmath}
\usepackage{footnote}
\usepackage{makecell}
\usepackage{multirow}
\restylefloat{table}
\usepackage{caption,subcaption}
\usepackage{tikz}
\usetikzlibrary{matrix}
\usepackage{blindtext}
\usepackage[export]{adjustbox}
\usepackage{cite}
\usepackage{amsmath}
\usepackage[font=small,skip=0pt]{caption}
\begin{document}

\title{Exploiting Multi-Scale Fusion, Spatial Attention and Patch Interaction Techniques for Text-Independent Writer Identification}
%\title{Writer Identification Using Multi-Scale Fusion, Spatial Attention and Patch Interaction Techniques}
%Writer Identification Using Different  Deep Learning Approaches
\author{Abhishek Srivastava\inst{1} \and Sukalpa Chanda\inst{2} \and 
Umapada Pal\inst{1}}
\authorrunning{ A. Srivastava et al.}
% First names are abbreviated in the running head.
% If there are more than two authors, 'et al.' is used.
%
\institute{Computer Vision \& Pattern Recognition Unit, Indian Statistical Institute, Kolkata, India\\
\email{abhisheksrivastava2397@gmail.com, umapada@isical.ac.in} \and
 Department of Computer Science and Communication, Østfold University College, Halden, Norway\\ 
\email{sukalpa@ieee.org}}
%\email{lncs@springer.com}\\
% \url{http://www.springer.com/gp/computer-science/lncs} \and
% ABC Institute, Rupert-Karls-University Heidelberg, Heidelberg, Germany\\
 %\email{\{abc,lncs\}@uni-heidelberg.de}}
%
\maketitle              % typeset the header of the contribution
\begin{abstract}
\textit{
  Text independent writer identification is a challenging problem that differentiates between different handwriting styles to decide the author of the handwritten text. Earlier writer identification relied on handcrafted features to reveal pieces of differences between writers. Recent work with the advent of convolutional neural network, deep learning-based methods have evolved. In this paper, three different deep learning techniques - spatial attention mechanism, multi-scale feature fusion and patch-based CNN were proposed to effectively capture the difference between each writer's handwriting. Our methods are based on the hypothesis that handwritten text images have specific spatial regions which are more unique to a writer's style, multi-scale features propagate characteristic features with respect to individual writers and patch-based features give more general and robust representations that helps to discriminate handwriting from different writers. The proposed methods outperforms various state-of-the-art methodologies on word-level and page-level writer identification methods on three publicly available datasets - CVL, Firemaker, CERUG-EN datasets and give comparable performance on the IAM dataset. 
}

\end{abstract}

\section{Introduction}
Handwriting of an individual is unique and this particular
phenomenon has been utilized by forensic handwriting experts for many decades. Handwriting experts today are aided by computer programs which actually can identify an individual on the basis of his handwriting, this  technique of identifying a writer from a document image using a software is termed as \enquote{Writer Identification}.  Over the last two decades many work has been published on  \enquote{Writer Identification}. But text independent Writer Identification in a limited data scenario is still a challenging task. It has found various applications in forensic~\cite{pervouchine2007extraction} and historical~\cite{brink2012writer} document analysis.
Before the advent of deep-learning techniques, handcrafted features like gradient, chain-code, allograph, texture etc., were mostly used for writer identification. These feature extraction techniques render  discriminating features in predicting the identity of the writer. Deep-learned features have shown impressive performance  in various types of image classification problem and \enquote{Writer Identification} is also not an exception. Deep learning-based methods in general demand a huge amount of annotated text for proper training. For an application like \enquote{Writer Identification} it might not be possible always to procure enough annotated data.  Over the last few years, some deep-learning based methods \cite{nguyen2019text}, \cite{javidi2020deep} have explored writer identification. To tackle these issues methods which require limited data for identification of the authors are required. Word level writer identification is challenging since very limited information about writer's pattern and technique is available to make a decision. Few deep learning based methodologies are available, for example, He et al.~\cite{he2020fragnet} proposed fragment based deep neural network to use convolution neural networks (CNN) for writer identification. CNN were able to learn high level features of the text block and recognize various discriminative features in the word image. CNN's have been previously used to capture local features at the sub-region and character level and combining them for writer identification. Attention based mechanisms are well suited to identify characteristic and discriminative region in an image and enhance the performance of visual recognition based systems. In case of text independent writer identification, the word image is constituted of various segments which capture the unique style of the person's handwriting. Previous deep learning methodologies fail to exploit the contribution of more informative regions of the text image. Recent advances in computer vision has generated interest in fusion of multi-scale features to obtain diverse and rich feature representations~\cite{wang2020deep}. Various resolution scales in handwritten text capture different aspects of a writer's style and structure of his/her handwriting, exploiting multi scale features and their fusion for eventual classification that obtains higher accuracy. \\

We devise three deep learning techniques to address and exploit those above mentioned facts and compare them to study the impact of different deep learning techniques for writer identification at word and page level. The contributions of our work are as follows:
\begin{itemize}
    \item  We propose a Spatial Attention network (SA-Net) which incorporates spatial attention to enhance relevant and informative feature maps and suppress irrelevant features for effective writer identification performance. Another potential discriminative features in text images are multi-scale features. 
    \item To achieve efficient multi-scale fusion, we customized the MSRF-Net~\cite{srivastava2021msrf} to a classification network suitable for writer identification.
    \item Inspired by He and Schomaker~\cite{he2020fragnet} we propose another patch based CNN named PatchNet which has separate pathways for each patch and uses a Dual Patch Dense Feature Exchange (DPDFE) block to exchange information across various patches, and making separate writer identity prediction for each patch.
    \item We attained new benchmarks on CVL, Firemaker and CERUG-EN datasets on word-level and page-level writer identification tasks.
\end{itemize}
The structure of this paper is as follows. Section~\ref{sec:relatedwork} provides an overview of related methods and strategies introduced over previous years. Section~\ref{sec:method} introduces our proposed methodologies for text independent writer identification. The details of our experiment settings and datasets used are presented in Section~\ref{sec:experiments}. In Section~\ref{section:resultsdiss} we report the results attained by our methods and their comparison with other state-of-the-art methods on word-level and page-level writer identification tasks, we conclude our paper in Section~\ref{sec:conclusion}.
\section{Related Work}
\label{sec:relatedwork}
The initial works in the field of writer identification were guided by handcrafted feature generation and later with the advent of deep-learning, deep-learning based writer identification methods were proposed. Before the deep-learning methods a wide variety of classifiers like SVM, K-NN, Neural Network were used along with different tools like PCA and LDA to magnify the discrminativeness of various hand crafted features. In the following two subsections we will have a brief discussion on handcrafted features for writer identification followed by deep-learning based approaches. 
\subsection{Hand Crafted Feature Based Writer Identification}
Difference in visual shapes in handwritten characters   has been exploited by considering  Connected component contour shapes, textural and allograph level features  in \cite{bulacu2007text}, Schomaker and Bulacu~\cite{schomaker2004automatic} proposed connected-component contours and its probability density function for writer identification.
%Namboodiri et al.~\cite{namboodiri2006text} identified repetitive primitives in handwritten texts and then calculated the similarity measures between the primitives for writer identification.%
Bulacu et al.~\cite{bulacu2007text} exploited to identify the writer. 
%Li et al.~\cite{li2007online} used stroke’s probability distribution function (SPDF) as writer specific features. 
 He et al.~\cite{he2008writer} used Hidden Markov Tree (HMT) in wavelet domain for writer identification. Tan et al.~\cite{tan2009automatic} developed a Continuous Character Prototype Distribution feature extraction technique and made classification using Minimum Distance method.
%Leng and Shamshuddin~\cite{leng2010writer} used Gray Scale Co-occurrence Matrices and multi-channel Gabor filtering for feature extraction. They used classifiers such as K-Nearest Neighbor (K-NN), Weighted Euclidean Distance, K-Means clustering and Kohen Self-Organizing Feature Map and compared both post-discretized and pre-discretized data. 
Jain and Doermann~\cite{jain2011offline} used K adjacent segments(KAS) to model character contours. The KAS features were clustered using a technique called affinity propagation to build a codebook for the bag of features model. 
%Awaida and Mahmoud~\cite{awaida2012state} combined component features, gradient distribution features, windowed gradient distribution features, contour chain code distribution features and windowed contour chain code distribution features and used classifiers such as Nearest Neighbor, Euclidean distance, Multiple Discriminant Analysis, Principal Component Analysis, Linear Discriminant Analysis, and Multidimensional Scaling. %
Jain and Doerman~\cite{jain2013writer} captured shape and curvature using contour gradients and used psuedo alphabets as features. Then writer identification was performed using K-Nearest Neighbour classifier.
He et al.~\cite{he2015junction} extracted features such as junction detection, final junction refinement quill and hinge and linked it with a learned codebook to increase performance. Chahi et al.~\cite{chahi2020cross} used connected components of the sub-images to extract features referred to as Cross multi-scale Locally encoded Gradient Patterns (CLGP). These CLGP histogram feature vectors were fed into a Nearest Neighbor classifier for writer identification.

%AI-Maadeed et al.~\cite{al2016novel} extracted features using edge detection probability distribution technique and classified using K-Nearest Neighbour method. 
%

\subsection{Deep Learning Feature Based Writer Identification}
Recently deep learning has drawn attention as convolutional neural networks(CNN) have proven effective in extracting discriminative features from handwritten texts. Initially, Fiel and Sablatnig~\cite{fiel2015writer} trained a CNN classifier and used the output of second last fully connected layer as features to perform nearest neighbour classification. Tang and Wu~\cite{tang2016text} performed data augmentation on handwritten documents to allow training of a deep CNN. The CNN is then used for feature extraction and Joint Bayesian technique is used for writer identification. DeepWriter~\cite{xing2016deepwriter} used multi-stream CNNs to learn diverse representation of text images. % Ni et al.~\cite{ni2017writer} explored writer identification on noisy handwritten document images using deep CNN for image denoising and extraction of features.%
Rehman et al.~\cite{rehman2019automatic} augmented text images using various techniques like contour, negatives and sharpness using text line images.
Multiple patches were generated from the text images and fed into an architecture similar to AlexNet pretrained on Imagenet to generate features. These features were classified using a support vector machine classifier. Keglevic et al.~\cite{keglevic2018learning} designed a triplet network to calculate similarity measure between different patches, and trained it by maximizing inter-class distance and minimizing intra-class distance. Global features of document is then calculated by aggregating vector of local image patch descriptors. Nguyen et al.~\cite{nguyen2019text} generated tuples of text images by randomly sampling characters as input for their CNNs. They trained CNNs to extract sub-region, character and global level features and effectively aggregated them to predict the identity of writer. He et al.~\cite{he2020fragnet} designed FragNet which first builds a global feature pyramid and then a local fragment pathway which leverages fragments of global feature pyramids to make separate writer identity prediction for each writer. Javidi and Jampour~\cite{javidi2020deep} quantified the thickness of handwritten documents using handwriting thickness descriptors(HTD). Resnet-18 was used to extract features from the text images and they were combined with HTDs for classification. In this work, we propose three different deep learning models which uses different architecture based components suitable for identifying and capturing various aspects of a writer's technique and style. 

\section{Methodology}
\label{sec:method}
In this section we discuss about our proposed approaches. We have developed the following methods.
\begin{enumerate}
    \item We develop a spatial attention based mechanism for identifying various author specific features of the word image. The characteristic style and features of the word occupy a very limited region in the word image. Generating a spatial attention map can help enhancing the features exploited from such regions. This serves as the basis of our spatial attention network(SA-Net) for writer identification.
    \item Multi-Scale features can capture information of varying spatial and receptive field sizes. The word images can have key discriminative features of diverse scale sizes which convey various characteristic features of writer. Thus, it is advantageous to design a writer identification system which effectively leverage multi-scale features while predicting the identity of our writer. We convert our MSRF-Net~\cite{srivastava2021msrf} to MSRF(Multi-Scale Residual Fusion) Classification network to effectively fuse multi-scale features and leverage them into predicting the identity of the writer more accurately. 
    \item Inspired by FragNet~\cite{he2020fragnet} we develop a patch based convolutional neural network called PatchNet. We use a different stream for each patch used and densely exchange various patch features using our Dual Patch Dense Feature Exchange (DPDFE) blocks. Each local patch predictions are then averaged over to make our final writer identity prediction.
\end{enumerate}
This section is structured as follows. In Section~\ref{section:spatial} we describe our spatial attention network(SA-Net), Section~\ref{section:MSRF} describes how we amend our MSRF network to a classification network and exploit multi-scale features of word images to develop a more accurate system for writer identification. Finally, in Section~\ref{section:Patch} we describe our proposed Patch-Net.

\subsection{Spatial Attention Network}
\label{section:spatial}
\begin{figure*}[!t]
    \centering
    \includegraphics[height=110pt,width=0.85\textwidth]{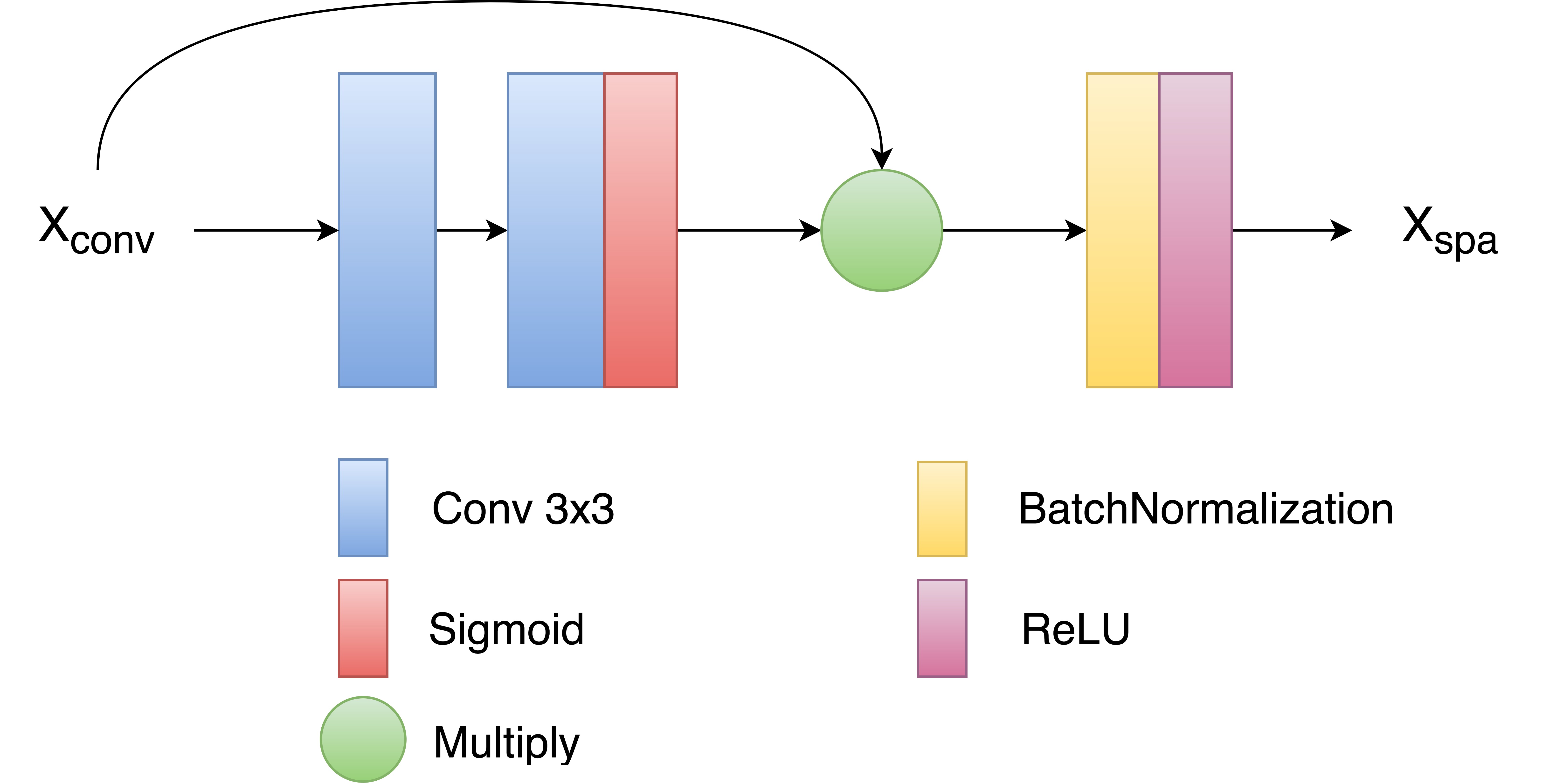}
    \caption{Architecture of Spatial attention unit employed in SA-Net}
    \label{fig:spatialnet}
\end{figure*}

In this section we introduce our spatial network for writer identification.
Specific regions of word images have characteristic textural and shape information which is unique to a specific writer. Characters in the word images also have a unique style in the manner they are written. To allow the identification and recognition of these regions we develop a spatial attention mechanism. Let  $I_{w}$ denote word images where ( $I_{w}\, \in \, R^{W\:\times\:H}$). The framework resembles a VGG-style network~\cite{simonyan2014very} where each $I_{w}$ is initially processed by a convolutional block.
Each convolutional block has 2 consecutive convolutional layers with 3 x 3 kernel size followed by batch-normalization and ReLU activation. This is described in Equation~\ref{eq:conv} where $X$ denotes the input tensor.
\begin{equation}{\label{eq:conv}}
    X_{conv} =  ReLU (BN(Conv(Conv(X))))
\end{equation}
The convolutional blocks are followed by a spatial attention unit (see Figure~\ref{fig:spatialnet}). This block comprises of two convolutional layers followed by a sigmoid activation function which calculates attention coefficient for each spatial location in the feature maps (see Equation~\ref{eq:spatial}). These attention maps are denoted as $A_{att}$. We multiply these attention maps described in Equation~\ref{eq:att} to suppress regions which are non relevant and enhance the spatial location of relevant and important feature maps.
\begin{equation}{\label{eq:spatial}}
    A_{att} = \sigma(Conv(X_{conv})) 
\end{equation}
\begin{equation}{\label{eq:att}}
    X_{spa} = X_{conv}\otimes A_{att} 
\end{equation}
$X_{spa}$ denotes the spatial attention enhanced feature maps which are then halved using max pooling. The number of feature maps in a convolutional and spatial attention unit are set to [64,128,256,512] respectively. We use adaptive average pooling at the last layer and a fully connected layer to make the final prediction or writer identity. For page level prediction we make predictions for all word images in the page and average over them as described in Equation~\ref{eq:spatialpage}, where N are the total word images in a page and $ID_{page}$ represents the identity of the writer. 
\begin{equation}{\label{eq:spatialpage}} 
    ID_{page} =  \frac{1}{N}\sum_{n=1}^{n=N} P(I_{w}^n)
\end{equation}

\subsection{MSRF Classification Network}
\label{section:MSRF}
\begin{figure*}[!t]
    \centering
    \includegraphics[height=180pt,width=0.9\textwidth]{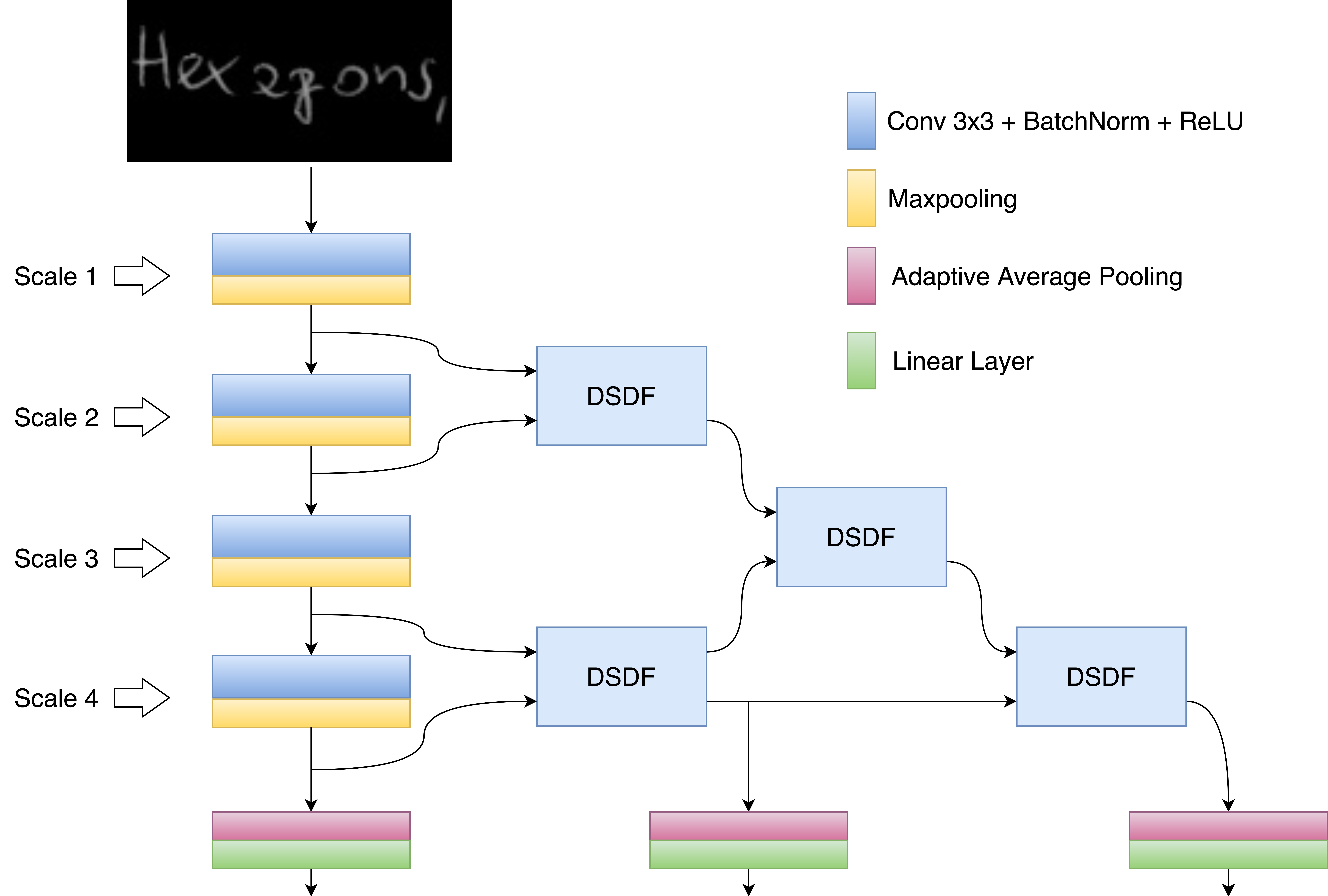}
    \caption{Architecture of Multi-scale Residual Fusion Classification Network}
    \label{fig:msrf}
\end{figure*}

Multi-scale feature exchange has been studied in past years in the field of computer vision. Fusion of multi-scale features result in diverse representations consequently generating richer and accurate feature maps. The word images are also structured such that different scale features capture varying writer characteristics. We use this motivation to convert our MSRF-Net~\cite{srivastava2021msrf} into a classification network (see Figure~\ref{fig:msrf}). Dual scale dense fusion (DSDF) blocks used in MSRF-Net serves the purpose of fusion of two different scaled features. The dense nature of the blocks allows features of various receptive fields to be generated and the residual connections allow relevant high-level and low-level features to be maintained while making final predictions. We modify the MSRF-Sub-network to translate it into a classification head. Contrary to the MSRF sub-network which aimed to fuse and exchange multi-scale features across all scales, we ensure that all different scaled representations are able to flow in the last scale level of the classification network (see Figure~\ref{section:MSRF}). To improve gradient flow, we allow last scale level of the MSRF classification network to make prediction before and after each DSDF block in the last scale level as shown in Figure~\ref{fig:msrf}. We use an adaptive pooling module and a fully connected layer in succession to make predict writer of the word image. Finally we average over all the predictions of to make our final predictions as shown in Equation~\ref{eq:msrfword}, where $C$ represents the number of classification layers in the MSRF classification network and $ID_{word}$ represents the identity of the writer for the word image.
\begin{equation}{\label{eq:msrfword}} 
    ID_{word} =  \frac{1}{C}\sum_{k=1}^{k=C} P(I_{w}^k)
\end{equation}
In order to make page level prediction, we again average over all word level predictions contained in the page as shown in Equation~\ref{eq:msrfpage}.
\begin{equation}{\label{eq:msrfpage}} 
    ID_{page} =  \frac{1}{N}\sum_{n=1}^{n=N} P(I_{w}^n)
\end{equation}

\subsection{PatchNet}
\label{section:patchnet}
\begin{figure*}[!t]
    \centering
    \includegraphics[height=180pt,width=0.9\textwidth]{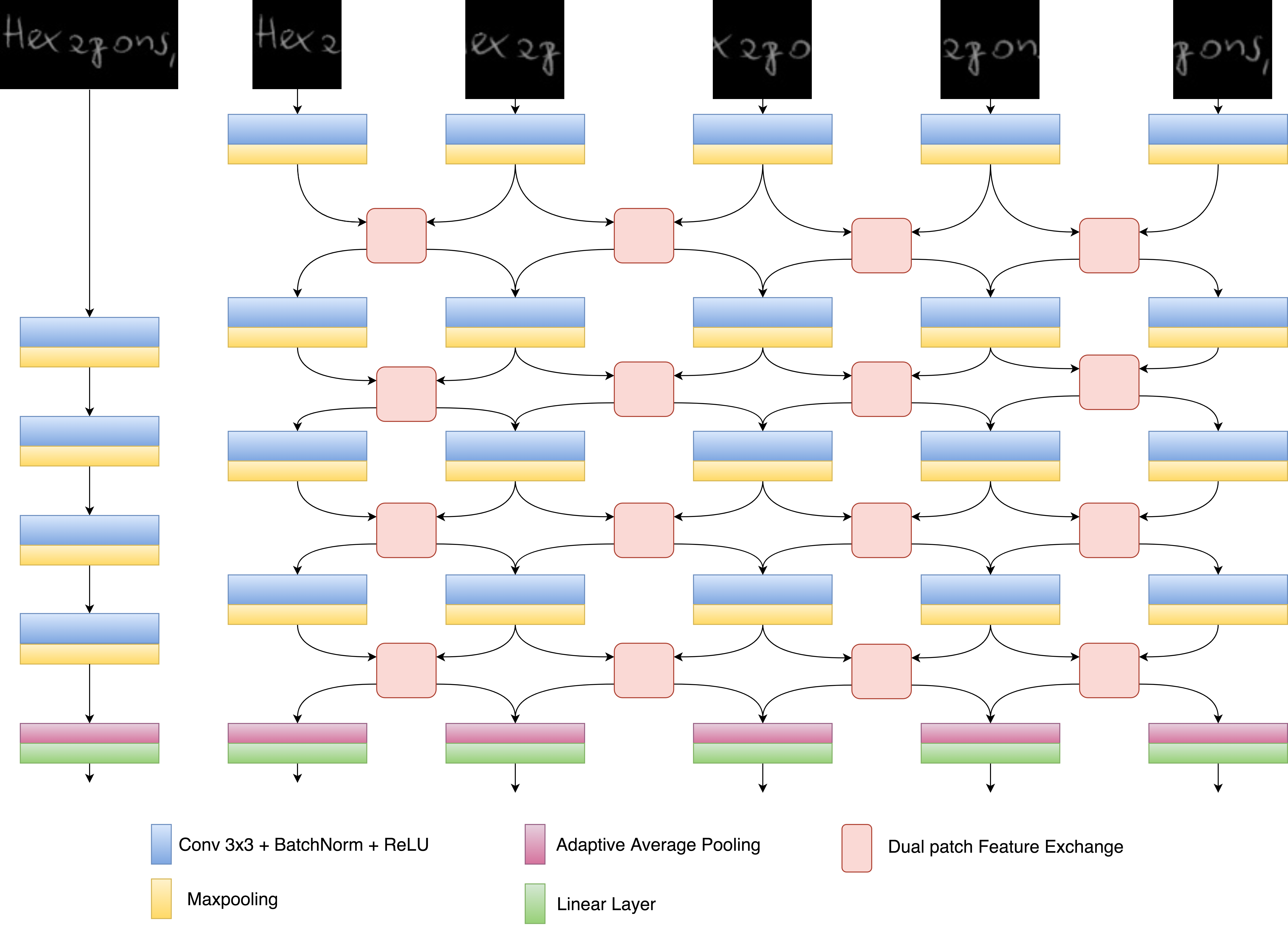}
    \caption{Architecture of our proposed PatchNet}
    \label{fig:patchRRDB}
\end{figure*}

Inspired by FragNet~\cite{he2020fragnet} we develop a patch based classification network (see Figure~\ref{fig:patchRRDB}) . The $\mathcal{I}_{w}$ is divided into patches of size 64 x 64. We generate 5 patches from the original 64 x 128 $\mathcal{I}_{w}$ and make different pathways for each patch. Each path has a initial convolutional unit of two successive convolutional layers, batch-normalization and ReLU activation. Which is followed by a maxpooling layer to reduce the spatial dimension by a factor of 2. To exchange information between two patches we design dual patch feature exchange (DPDFE) block. The entire convolutional unit, DPDFE blocks and max-pooling sequence is repeated 4 times to make patch level predictions.
We also use a global prediction pathway which has a similar architecture as SA-Net without the spatial attention unit.
Each patch level predictions and global prediction are averaged to make the final prediction. Page level predictions are made according to Equation~\ref{eq:patchpage}.
\begin{equation}{\label{eq:patchpage}} 
    ID_{page} =  \frac{1}{N}\sum_{n=1}^{n=N} P(I_{w}^n)
\end{equation}
\subsubsection{Dual patch Dense feature exchange Blocks}
\label{section:Patch}
\begin{figure*}[!t]
    \centering
    \includegraphics[height=220pt,width=0.9\textwidth]{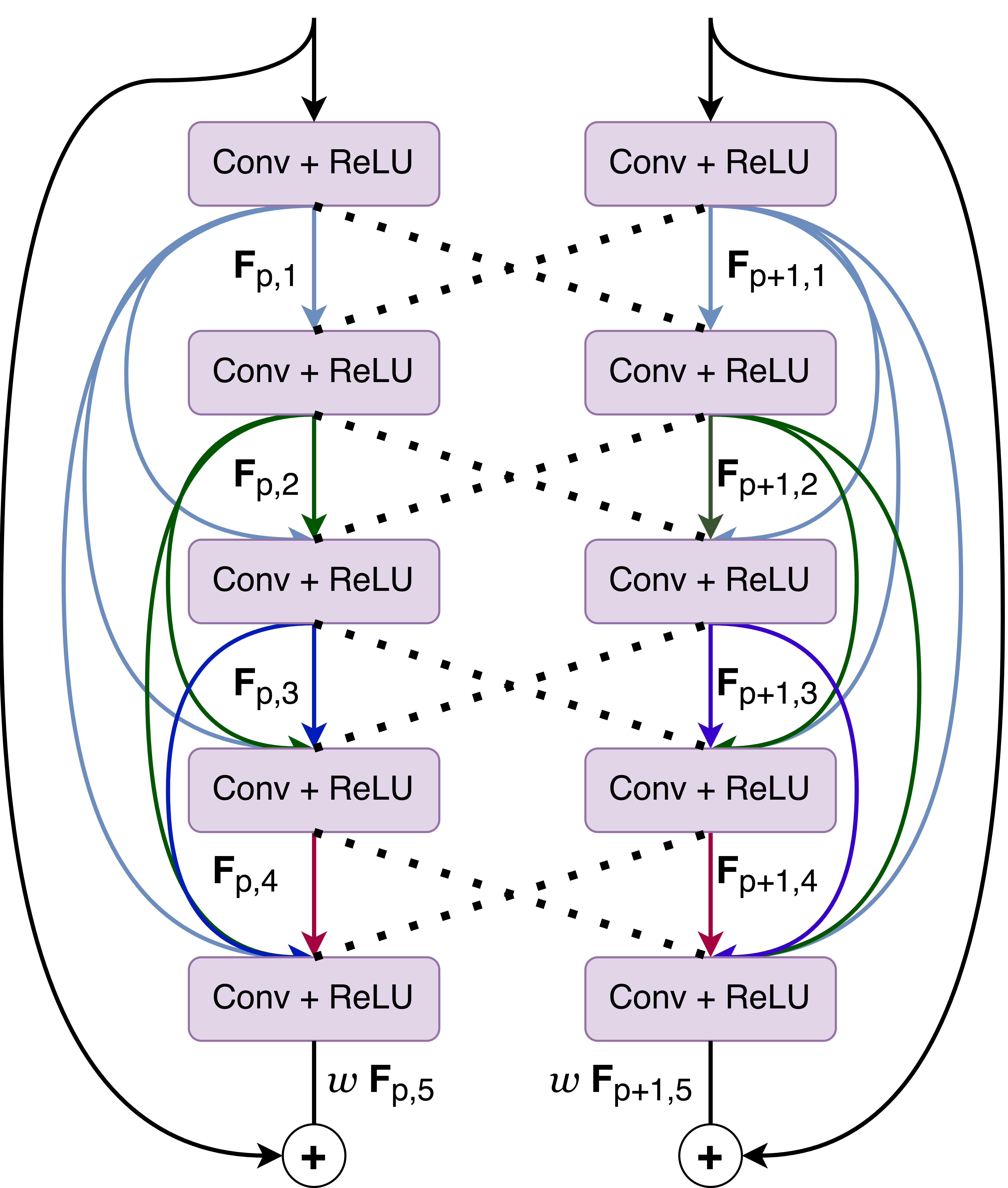}
    \caption{Architecture of Dual Patch Dense Feature Exchange Block(Dotted lines represent features incoming from parallel patch stream)}
    \label{fig:dpfe}
\end{figure*}

In this section, we describe the structure of our dual patch dense feature exchange blocks.
Let two successive patches be denoted by $\mathcal{I}_{p}$ and  $\mathcal{I}_{p+1}$. The feature maps generated by convolutional unit of each patch stream be denoted by $M_{p,l}$, where $l$ denotes how many layers of DPDFE blocks the feature maps have been processed by and initially $l=0$. The DPDFE blocks are residual dense blocks which takes feature maps of two different patches and process each of them using two different densely connected streams (see Figure~\ref{fig:dpfe}). Each stream has 5 densely connected convolutional layers, Let the output of each such layer be $F_{p,c}$ where $p$ denotes which patch is being processed and c denotes which convolutional layer has processed the feature maps in the dense stream. After each convolutional layer in the dense stream, the two different patch streams exchanges features as described in Equation~\ref{eq:featuepatch}($M_{p,l}$ and $F_{p,0}$ are the same).
\begin{equation}{\label{eq:featuepatch}}
    M_{p,l+1} = F_{p,c} \oplus F_{p,c-1} \oplus F_{p,c-2}\oplus \cdots \oplus F_{p+1,0}
\end{equation}

We again scale the output features of DPDFE blocks by a factor of $w=0.4$ to avoid instability~\cite{lim2017enhanced,szegedy2017inception} and add it back to the input of the respective DPDFE block as shown in Equation~\ref{eq:5}.
\begin{equation}{\label{eq:5}}
M_{p,l+1} = w \times {M_{p,l+1}} + M_{p,l}
\end{equation}
\section{Experiments}
\label{sec:experiments}
In this section we describe the writer identification datasets used for our experiments. We also describe the implementation details of our three deep learning-based writer identification methods. We use the training and testing split used by He and Schomaker in FragNet~\cite{he2020fragnet}. It is ensured that each word image from each page either occurs in the training split or in the testing split, which makes the methods suitable for both word-level and page-level writer identification tasks.
\subsection{Datasets}
We benchmark our methods on four publicly available datasets namely: CERUG-EN~\cite{he2015junction}, Firemaker~\cite{schomaker2000forensic}, CVL~\cite{kleber2013cvl} and IAM~\cite{marti2002iam}
\begin{enumerate}
    \item CERUG-EN~\cite{he2015junction} has 105 documents, predominantly from Chinese students. There are two paragraphs in English where one paragraph is used for training and another is used for testing. Since word images are not provided separately we use the roughly segmented word images provided by He and Schomaker in the publicly released code of FragNet.
    \item Firemaker~\cite{schomaker2000forensic} has 250 different writers where each writer writes four pages. First page is used for training and the fourth page is used for testing.
    \item CVL~\cite{kleber2013cvl} has 310 writers. Each person has written five pages of text with 27 writers contributing seven pages. First three pages are used for training and the rest are used for testing.
    \item IAM~\cite{marti2002iam} has 610 different writers contributing varying amount of text. When more than one page is available for a writer, we choose one page for training and rest for testing. When only one page is available the lines are divided into training and testing subsets. The word images are publicly available.

\subsection{Implementation details}
We pre-process the images to 64 x 128 while maintaining aspect ratio. To avoid distortion white pixel padding is done. We use a batch size of 16 and train all methods for 50 epochs. We follow the training setting of FragNet to ensure fair and complete comparison. Adam optimizer is used with intial learning rate 0.0001 and a weight decay of $1e-4$. We decay the learning rate by a factor of 0.5 after every 10 epochs. The feature maps in each level of classification networks are [64,128,256,512] for all 3 methods.
The FLOPs of WordImgNet are 1.05G, whereas FLOPs of FragNet-64, FragNet-32, Frag-Net-16 are approximately 7.14G, 7.41G and 3.90G, respectively.
The FLOPs of MSRF-Classification network, SA-Net and PatchNet, and ResNet18+HTDs are around 5.5G, 4.10G, and 7.65G respectively. The proposed models are available at \href{https://github.com/NoviceMAn-prog/SA-Net-MSRF-CNet-and-PatchNet-for-Writer-Identification}{https://github.com/NoviceMAn-prog/SA-Net-MSRF-CNet-and-PatchNet-for-Writer-Identification}.
% MSRF - 5,496,024,499 FLOPs or approx. 5.50 GFLOPs
% Spatial - 4,104,299,008 FLOPs or approx. 4.10 GFLOPs
% PATCH RRDB- 7,652,636,262 FLOPs or approx. 7.65 GFLOPs

\end{enumerate}
\section{Results \& Discussion}
\label{section:resultsdiss}
In this section we will compare our MSRF Classification Network, SA-Net and PatchNet with other published state-of-the-art methods on word-level and page-level writer identification task. It is worth mentioning here that  there exists many deep-learning methods for writer identification, and even though those experiments were conducted on public datasets, lack of publicly available source code of those published methods  creates hindrance towards a fair comparison. Additionally we chose methods that were designed for word level writer identification. Keeping those factors in mind, we could compare our methods  with \cite{he2020fragnet}, \cite{javidi2020deep} as those methods have released their code. We train and test those methods using the same set of training and test images as we did for our proposed methods for an unbiased comparison. We establish new state-of-the-art writer identification results on three benchmark datasets - CVL, Firemaker and CERUG-EN datasets.  Section~\ref{section:otherworks} describes various other state-of-the-art deep learning methods we select for comparison with our methods on word-level and page-level writer identification tasks. In Section~\ref{section:wordlevel} we provide quantitative comparison of our methods with other baselines on word level writer identification tasks. Section~\ref{section:pagelevel} provides writer identification results on page level writer identification task.
\subsection{Comparison with Other Published Methods}
\label{section:otherworks}
To provide exhaustive comparison of our methods with other baselines we select ResNet18~\cite{he2016deep},ResNet18 conjugated with handwriting thickness descriptors(HTD)~\cite{javidi2020deep}, WordImgNet~\cite{he2020fragnet} and FragNet-q~\cite{he2020fragnet} where q represents the q x q fragment size. 
\begin{enumerate}
    \item ResNet18 is a standard computer vision classification baseline.
    \item FragNet is fragmentation based CNN with two streams. First global feature pyramid used for extraction of features. Second stream is a fragment pathway to process fragments of the original image and receive fragments from the global feature pyramid to make prediction. Each fragment has its own prediction and the final prediction is made by averaging over all local fragment predictions.
    We use FragNet-64, FragNet-32 and FragNet-16 for our experiments.
    \item WordImgNet is designed such that the entire image is fed into a CNN framework identical to the fragment pathway to make a single global prediction.
    \item ResNet18 + HTDs, ResNet18 captures high level features of the input text image. HTDs are spatial descriptor that analyze a writer's handwriting thickness depending upon factors like pressure of pen, unique style. The features extracted by ResNet18 are concatenated  with HTDs which serves as additional discriminative features.
\end{enumerate}
\subsubsection{Result on Word level Writer Identification}
\label{section:wordlevel}
In this section we compare the Top-1 and Top-5  writer identification accuracy of our proposed methods with other state-of-the-art methods at word level. In Table~\ref{tab:resultword} we present the detailed comparison of all methods on all four datasets. In CERUG-EN~\cite{he2015junction} we observe that SA-Net gives the best performance outperforming the previous state-of-the-art performer FragNet-64 by 4.7\% accuracy in Top-1 and by 1.5\% in Top-5. We can notice that along with SA-Net, MSRF Classification also beats FragNet-64 in performance in both Top-1 and Top-5 accuracy, PatchNet is comparable to it in performance. For Firemaker dataset, our SA-Net again gives the best performance gaining 3.2\% and 0.8\% in Top-1 and Top-5 accuracy over FragNet-64. MSRF-Net gives the second best performance gaining 2.2\% and 0.8\% in Top-1 and Top-5 accuracy over FragNet. In CVL dataset writer identification problem, MSRF Classification obtains the best performance achieving 91.4\% Top-1 and 97.6\% Top-5 accuracy outperforming FragNet-64 by 1.2\% and 0.1\% in Top-1 and Top-5 accuracy. SA-Net also performs better than FragNet-64 achieving 90.7\% and 97.4\% Top-1 and Top-5 accuracy. PatchNet gives a comparable 86.1\% Top-1 and 96.3\% Top-1 and Top-5 accuracy respectively. On the writer identification task on IAM dataset, FragNet-64 reports the best Top-1 accuracy of 85.1\% while the best Top-5 accuracy is shared between FragNet-64 and MSRF Classification network, both achieving 95\%. The superior performance of SA-Net on two datasets i.e. Firemaker and CERUG-EN shows the potential of spatial attention mechanism's ability to extract relevant differentiating elements of a writer's handwriting . The multi scale features obtained and fused in MSRF classification network obtains the highest Top-1 and Top-5 accuracy on CVL dataset. This displays the capacity of multi-scale features to identify the characteristics of writer's style in  his handwriting. Although PatchNet outperforms previous state-of-the-art methods on only one dataset, it shows the potential of patch or fragment based networks for writer identification.

\begin{table}[!t]
\centering
%\scriptsize
\footnotesize
\caption{Result comparison on word level writer identification}
\begin{tabular}{@{}l|l|l|l|l|l|l|l|l@{}}
\toprule
\multirow{2}{*}{\textbf{Method}} & \multicolumn{2}{|c|}{IAM} & \multicolumn{2}{|c|}{CVL} & \multicolumn{2}{|c|}{Firemaker} & \multicolumn{2}{|c|}{CERUG-EN} \\
& Top-1 & Top-5 & Top-1 & Top-5 & Top-1 & Top-5 & Top-1 & Top-5 \\ \hline
ResNet18~\cite{he2016deep} & 83.2 & 94.3 & 88.5 & 96.7 & 63.9 & 86.4 & 70.6 & 94.0 \\ \hline
ResNet18+HTD~\cite{javidi2020deep} & 76.9 & 91.6 & 85.1 & 95.6 & 60.7 & 82.6 & 70.1 & 91.8 \\ \hline
WordImgNet~\cite{he2020fragnet} & 81.8 & 94.1 & 88.6 & 96.8 & 67.9 & 88.1 & 77.3 & 96.4 \\ \hline 
FragNet-16~\cite{he2020fragnet} & 79.8 & 93.3 & 89.0 & 97.2 & 59.6 & 83.2 & 60.6 & 90.3 \\ \hline
FragNet-32~\cite{he2020fragnet} & 83.6 & 94.8 & 89.0 & 97.3 & 65.0 & 86.8 & 62.3 & 90.1 \\ \hline
FragNet-64~\cite{he2020fragnet} & \textbf{85.1} & \textbf{95.0} & 90.2 & 97.5& 69.0 & 88.5 & 77.5 & 95.6 \\ \hline
Patch (Proposed) & 80.2 & 93.5 & 86.1 & 96.2 & 62.4 & 84.9 & 77.1 & 96.5 \\ \hline
SA-Net (Proposed)& 83.4 & 94.6 & 90.7 & 97.4 & \textbf{72.2} & \textbf{89.3} & \textbf{82.2} & \textbf{97.1} \\ \hline
MSRF-Net (Proposed)& 84.6 & \textbf{95.0} & \textbf{91.4} & \textbf{97.6} & 71.2 & \textbf{89.3} & 79.6 & 96.8 \\ \hline

\bottomrule
\end{tabular}
\label{tab:resultword}
\end{table}

\subsubsection{Result on Page level Writer Identification}
\label{section:pagelevel}
In this section we provide the quantitative analysis of the comparison between our proposed methods and other state-of-the-art methods on page level writer identification. In the Firemaker dataset writer identification task, the proposed SA-Net outperforms FragNet-64 by 0.4\% in Top-1 accuracy. SA-Net reports a Top-5 accuracy of 99.6\% which is equal to the Top-5 accuracy to FragNet-64. SA-Net reports the highest Top-1 page level accuracy on CERUG-EN of 99.1\%. Additionally, FragNet-64, PatchNet, SA-Net,MSRF classification network and ResNet18+HTDs all tie for the best Top-5 performance of 100\% on CERUG-EN. MSRF-Net and SA-Net both outperforms FragNet-64 by 0.3\% on Top-1 page level accuracy on the CVL dataset. MSRF-Net report the highest 99.6\% Top-5 page level accuracy while FragNet-64 and SA-Net gives 99.4\% Top-5 page level accuracy. For IAM dataset, FragNet-64 reports the highest 96.3\% Top-1 page level accuracy. SA-Net and MSRF-Net reports the first and second best Top-5 accuracy of 98.2\% and 98.1\%, respectively. We notice that again SA-Net and MSRF-Net attains new benchmarks on IAM, CVL, Firemaker and CERUG-EN datasets, exhibiting the potential of amplified features on the basis of spatial attention and multi-scale features.

\begin{table}[!t]
\centering
%\scriptsize
\footnotesize
\caption{Result comparison on page level writer identification}
\begin{tabular}{@{}l|l|l|l|l|l|l|l|l@{}}
\toprule
\multirow{2}{*}{\textbf{Method}} & \multicolumn{2}{|c|}{IAM} & \multicolumn{2}{|c|}{CVL} & \multicolumn{2}{|c|}{Firemaker} & \multicolumn{2}{|c|}{CERUG-EN} \\
& Top-1 & Top-5 & Top-1 & Top-5 & Top-1 & Top-5 & Top-1 & Top-5 \\ \hline
ResnNet18+HTD~\cite{javidi2020deep} & 95.2 & 98.0 & 98.3 & 98.3 & 98.0 & 99.2 & 98.0 & \textbf{100} \\ \hline
WordImgNet~\cite{he2020fragnet} & 95.8 & 98.0 & 98.8 & 99.4 & 97.6 & 98.8 & 97.1 & 100 \\ \hline 
FragNet-16~\cite{he2020fragnet} & 94.2 & 97.4 & 98.5 & 99.4 & 92.8 & 98.0 & 79.0 & 97.1 \\ \hline
FragNet-32~\cite{he2020fragnet} & 95.3 & 98.0 & 98.6 & 99.4 & 96.0 & 99.2 & 84.7 & 97.1 \\ \hline
FragNet-64~\cite{he2020fragnet} & \textbf{96.3} & 98.0 & 99.1 & 99.4 & 97.6 & \textbf{99.6} & 98.1 & \textbf{100} \\ \hline
Patch (Proposed) & 93.6 & 96.9 & 99.0 & 99.3 & 95.6 & 98.4 & 98.1 & \textbf{100} \\ \hline
SA-Net (Proposed) & 94.7 & \textbf{98.2} & \textbf{99.4} & 99.4 & \textbf{98.0} & \textbf{99.6} & \textbf{99.1} & \textbf{100} \\ \hline
MSRF-Net (Proposed) & 94.8 & 98.1 & \textbf{99.4} & \textbf{99.6} & 97.2 & 99.2 & 98.1 & \textbf{100} \\ \hline
\bottomrule
\end{tabular}
\label{tab:resultpage}
\end{table}

\section{Conclusion}
\label{sec:conclusion}
In this paper we proposed three deep learning based solutions for text-independent writer identification. Our proposed SA-Net was able to identify and enhance the feature flow from spatial regions more relevant and significant in determining the identity of the writer. MSRF Classification network performed multi-scale feature fusion to gather more diverse representations consisting of features having varying receptive fields. The residual nature of the dual scale dense fusion (DSDF) blocks allow an effective combination of high- and low-level feature representations to be available at the disposal of final classification layer to make more accurate predictions. On the other-hand, PatchNet allows effective feature exchange between different patch streams to make more robust predictions. Our methods were able to outperform previous state-of-the-art methods for word-level and page-level writer identification on CVL, Firemaker and CERUG-EN datasets, while giving comparable performance on the IAM dataset. We show that developing deep learning based mechanisms exploiting spatially relevant regions and multi scale features is also a viable option to increase performance of writer identification systems. 
\section{Acknowledgement}
This is a collaborative research work between Indian Statistical Institute, Kolkata, India and Østfold University College, Halden, Norway. The experiments in this paper were performed on a high performance computing platform \enquote{Experimental Infrastructure for Exploration of Exascale Computing} (eX3), which is funded by the Research Council of Norway.

\bibliographystyle{splncs04}
\bibliography{ms}
\end{document}